\documentclass[runningheads]{llncs}
\usepackage[utf8]{inputenc}
\usepackage[vietnamese, english]{babel}
\usepackage{graphicx}
\usepackage[colorlinks]{hyperref}

\begin{document}
\title{An Efficient Model for Sentiment Analysis of Electronic Product Reviews in Vietnamese \thanks{Supported by Kyanon Digital}}
\titlerunning{Sentiment Analysis of Electronic Product Reviews  in Vietnamese}
%
\author{Suong N. Hoang\inst{1, 2}\orcidID{0000-0002-3354-013X} \and
Linh V. Nguyen\inst{1}\orcidID{0000-0003-0776-9480} \and
Tai Huynh\inst{1,2} \and
Vuong T. Pham\inst{1,3}
}
\authorrunning{Suong et al.}
%
\institute{Kyanon Digital, Ho Chi Minh City, Vietnam\\
\email{\{suong.hoang,linh.nguyenviet,tai.huynh,vuong.pham\}@kyanon.digital} \\
\url{https://kyanon.digital} \and
Advosights, Ho Chi Minh City, Vietnam \\
\email{\{suong.hoang,tai.huynh\}@advosights.com}\\
\url{https://advosights.com} \and
Saigon University, Ho Chi Minh City, Vietnam \\
\email{vuong.pham@sgu.edu.vn}
}
\maketitle              
\begin{abstract}
In the past few years, the growth of e-commerce and digital marketing in Vietnam has generated a huge volume of opinionated data. Analyzing those data would provide enterprises with insight for better business decisions. In this work, as part of the Advosights project, we study sentiment analysis of product reviews in Vietnamese. The final solution is based on Self-attention neural networks, a flexible architecture for text classification task with about $90.16\%$ of accuracy in $0.0124$ second, a very fast inference time.
\end{abstract}
\section{Introduction}
Sentiment analysis aims to analyze human opinions, attitudes, and emotions. It has been applied in various fields of business. For instance, in our current project, Advosights, it is used to measure the impact of new products and ads campaigns through consumer's responses.

In the past few years, together with the rapid growth of e-commerce and digital marketing in Vietnam, a huge volume of written opinionated data in digital form has been created. As the result, sentiment analysis plays a more critical role in social listening than ever before. So far, human effort is the most common solution for sentiment analysis problems.
However, this approach generally does not result in the desired outcomes and speed. Human check and labeling are time consuming and error-prone. 
Therefore, developing a system that automatically classifies human sentiment is highly essential.

While we can easily find a lot of sentiment analysis researches for English, there are only a few works for Vietnamese. Vietnamese is a unique language and it differs from English in a number of ways. To apply the same techniques that work for English to Vietnamese would yield inaccurate results. This has motivated our systematic study in sentiment analysis for Vietnamese. Since our project, Advosights, initially served a well-known electronics brand, we decided to focus our study on electronic product reviews. Broader scopes will be studied in future works. 

Our initial approach was to build a sentiment lexicon dictionary. Its first version was based on some statistical methods \cite{esuli2006,baccianella2010,mohammad2013} to estimate the sentiment score for each word from a list, collected manually  based on Vietnamese dictionaries. This approach did not work well because the dataset came from casual reviews, that were practically spoken language with a lot of slang words and acronyms. This fact made it almost impossible to build a dictionary that cover all of those words. We then tried to use a simple neural network to learn sentiment lexicons from corpus automatically \cite{vo2016}. This also did not work well because some words in Vietnamese have same morphology, but they have different meanings in different contexts. For example, the words {\em ``đã"} in two sentences ``nhìn {\em đã} quá" and ``{\em đã} quá cũ" have different meanings. But by using the dictionary, they have the same sentiment score.

Some machine learning-based approaches have been studied. For examples, CountVectorizer and Term Frequency–Inverse Document Frequency (Tf-idf) were used for word representations. Support Vector Machine (SVM) and Naive Bayes were used as classifiers. However, the results were not very encouraging.


We also investigated various types of recurrent neural networks (RNNs) such as long short-term memory(LSTM) \cite{hochreiter1997}, Bi-Directional LSTM (biLSTM) \cite{schuster1997} or gated recurrent unit (GRU) \cite{chung2014}, etc. Although some of them achieved pretty good accuracy, the models were heavy and had very long inference time. Our final model is based on the Self-attention neural network architecture Transformer \cite{vaswani2017}, a well known state of the art technique in machine translation. It provided top accuracy and has very fast inference time when running on real data. 

The paper is organized as follows. In section~\ref{S:Background}, some description of self-attention is provided for motivation. In section~\ref{S:Approach}, our architecture is presented. The experiments are described in section~\ref{S:Expriment}. Finally, conclusions and remarks are included in section~\ref{S:conclusion}. 

\section{Background} \label{S:Background}
Inspired by human sight mechanism, Attention was used in the field of visual imaging about 20 years ago \cite{schneider1998}. In 2014, a group from Google DeepMind applied Attentions to the RNN for image classification tasks \cite{mnih2014}. After that, Bahdanau et al. \cite{bahdanau2014} applied this mechanism to encoder-decoder architectures in machine translation task. It became the first work to apply Attention mechanism to the field of Natural Language Processing (NLP). Since then, Attention became more and more common for the improvement in various NLP tasks based on neural networks such as RNN/CNN~\cite{cheng2016,lu2016,kokkinos2017,daniluk2017,zhou2018}.\\

In 2017, Vaswani et al. first introduced Self-attention Neural Network~\cite{vaswani2017}. The proposed architecture, Transformer, did not follow the well-known idea of recurrent network. This paper paved the way and Self-attention have become a hot topic in the field of NLP in the last few years. In this section, we describe their approach in detail. 
\subsection{Attention}
The first description of Attention Mechanism in Machine Neural Translation \cite{bahdanau2014} was well known as a process to compute weighted average context vectors for each state of the decoder $s_i$ by incorporating the relevant information from all of the encoder states $h_j$ with the previous decoder hidden state $s_{i-1}$, which is determined by a alignment weights $\alpha_{ij}$ between each encoder state and previous hidden state of the decoder, to predict next state of the decoder. It can be summarized by the following equations:
\begin{equation}
c_i = \sum_{j=1}^{n}\alpha_{ij}h_{j}
\end{equation}
\begin{equation}
\alpha_{ij} = \frac{exp(e_{ij})}{\sum_{k=1}^{n}exp(e_{jk})}
\end{equation}
$e_{ij} = a(s_{i-1}, h_{j})$, where $a(s_{i-1}, h_j)$ is a function to compute the compatibility score between $s_{i-1}$ and $h_j$.
\subsubsection{Scaled Dot-Product Attention:}
Let us consider $s_{i-1}$ as a query vector $q$. And $h_j$ now duplicated, one is key vector $k_j$ and the other is value vector $v_j$ (in current NLP work, the key and value vector are frequently the same, there for $h_j$ can be considered as $k_j$ or $v_j$). The equations outlined above generally look like:
\begin{equation}
c = \sum_{j=1}^{n}\alpha_{j}v_{j}
\end{equation}
\begin{equation}
\alpha_{j} = \frac{exp(e_{j})}{\sum_{k=1}^{n}exp(e_{k})}
\end{equation}

In \cite{vaswani2017} paper, Vaswani et al. using the scaled dot-product function for the compatibility score function
\begin{equation}
e_{j} = a(q,k_{j}) = \frac{qk_{j}^{T}}{\sqrt{d_{model}}}
\end{equation}
where $d_{model}$ is dimension of input vectors or $k$ vector ($q$, $k$, $v$ have the same dimension as input embedding vector).
\subsubsection{Self-attention:}
Self-attention is a mechanism to apply Scaled Dot-Product Attention to every token of the sentence for all others. It means for each token, this process will compute a context output that incorporates informations of itself and information about how it relates to others tokens in the sentence.\\

By using a linear feed-forward layer as a transformation to create three vectors (query, key, value) for every token in sentence, then apply the attention mechanism outlined above to get the context matrix. But it seems very slow and takes a bunch of time for whole process. So, instead of creating them individually, we consider $Q$ is a matrix containing all the query vectors $Q = [q_1, q_2,...,q_n]$, $K$ contains all keys $K = [k_1, k_2,...,k_n]$, and $V$ contains all values $V = [v_1, v_2,...,v_n]$. As the result, this process can be done in parallel \cite{vaswani2017}.
\begin{equation}
Attention(Q, K, V ) = softmax(\frac{QK^{T}}{\sqrt{d_{model}}})V
\end{equation}
\subsubsection{Multi-head Attention}
Instead of performing Self-attention a single time with $(Q, K, V)$ of dimensions $d_{model}$. Multi-head Attention performs attention $h$ times with $(Q,K,V)$ matrices of dimensions $d_{model}/h$, each time for applying Attention, it is called a head. For each head, the (Q,K,V) matrices are uniquely projected with different dimensions $d_q$, $d_k$ and $d_v$ (equal to $d_{model}/h$), then self-attention mechanism is performed to yield an output of the same dimension $d_{model}/h$ \cite{vaswani2017}. After all, outputs of $h$ heads are concatenated, and apply a linear projection layer once again. This process can be summarized by the following equations:
\begin{equation}
MultiHead(Q, K, V) = Concat(head_1, head_2,..., head_h)W^{O}
\end{equation}
\begin{center} where $head_i = Attention(QW^{Q}_{i}, KW^{K}_{i}, VW^{V}_{i})$ \end{center}
Where the projections are parameter matrices $W^Q_i \in \mathsf{R}^{d_{model}\times d_k}, W^K_i\in \mathsf{R}^{d_{model} \times d_k}, \\W^V_i\in \mathsf{R}^{d_{model}\times d_v}, W^O\in \mathsf{R}^{hd_v \times d_{model}}$.
\subsection{Positional Information Embedding Representation}
Self-attention can provide context matrix containing information about how a token relates with the others. However, this attention mechanism still has limit, losing positional information problem. It does not care about the order of tokens. That means outputs of this process is invariant with the same set of tokens with order permutations. So, to make it work, neural networks need to incorporate positional information to the inputs. Sinusoidal Positional Encoding technique is commonly used to solve this problem.
\subsubsection{Sinusoidal Position Encoding:}
This technique was proposed by Vaswani et al. \cite{vaswani2017}. The main point of this technique is to create Position Encoding ($PE$) using sinusoidal and cosinusoidal functions to encode the position. The $PE$ function can be write by following equation:
\begin{equation}
PE(position,2i) = Sin(\frac{position}{10000^{2i/d_{model}}})
\end{equation}
\begin{equation}
PE(position,2i+1) = Cos(\frac{position}{10000^{2i/d_{model}}})
\end{equation}
where $position$ starts from $1$ and $i$ is $i_{th}$ dimension of $d_{model}$ dimensions. It means that for each dimension of the positional encoding corresponds to a different sinusoids.

The advantages of this technique is it can add positional information for sentences longer than those in training dataset.
\section{Our Approach} \label{S:Approach}
\subsection{Model architecture}
\begin{figure}[!htbp]
\centering
\begin{minipage}{.5\textwidth}
\centering
\includegraphics[width=\textwidth]{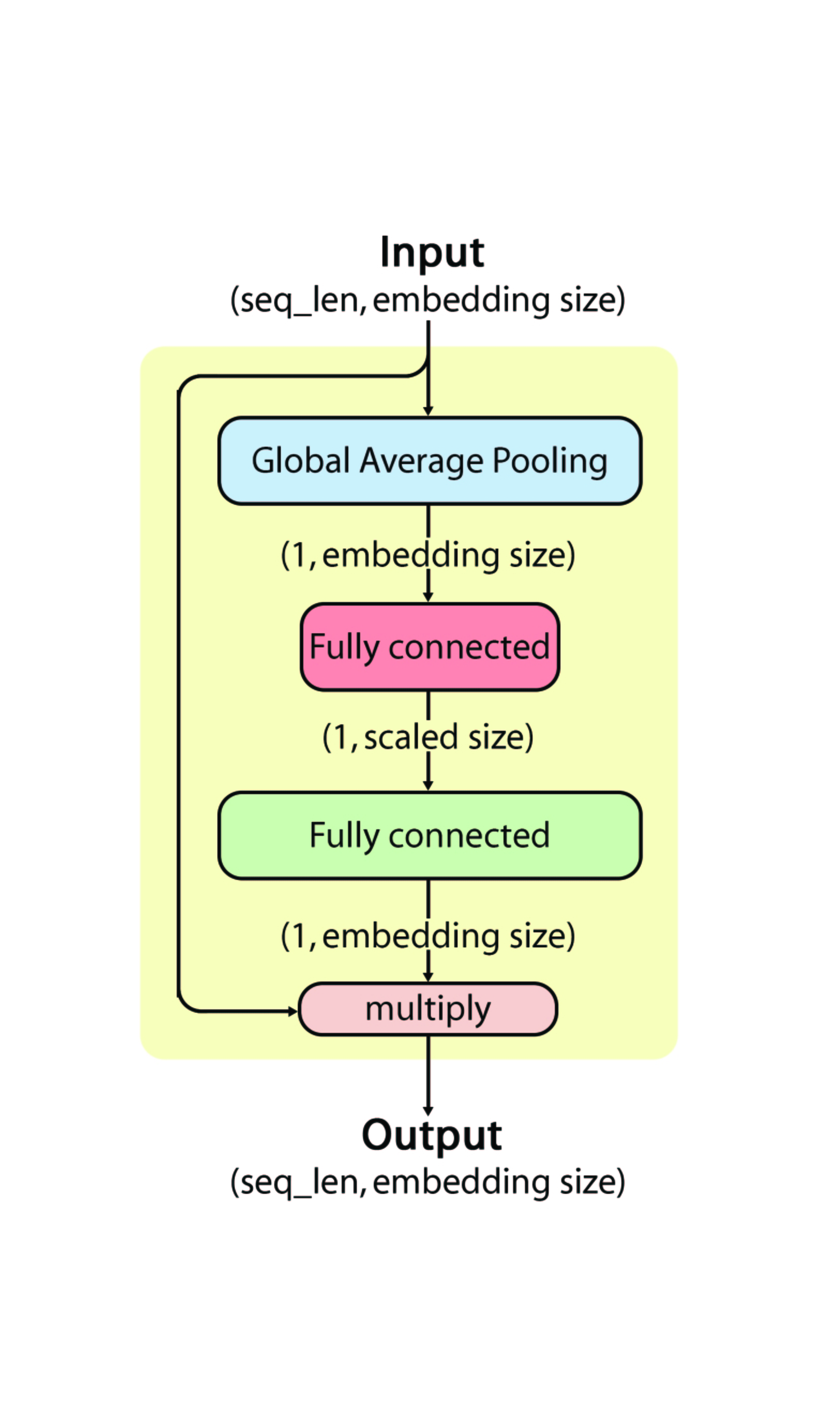}
\caption{Squeeze-Excitation architecure.} \label{fig1}
\end{minipage}%
\begin{minipage}{0.5\textwidth}
\centering
\includegraphics[width=\textwidth]{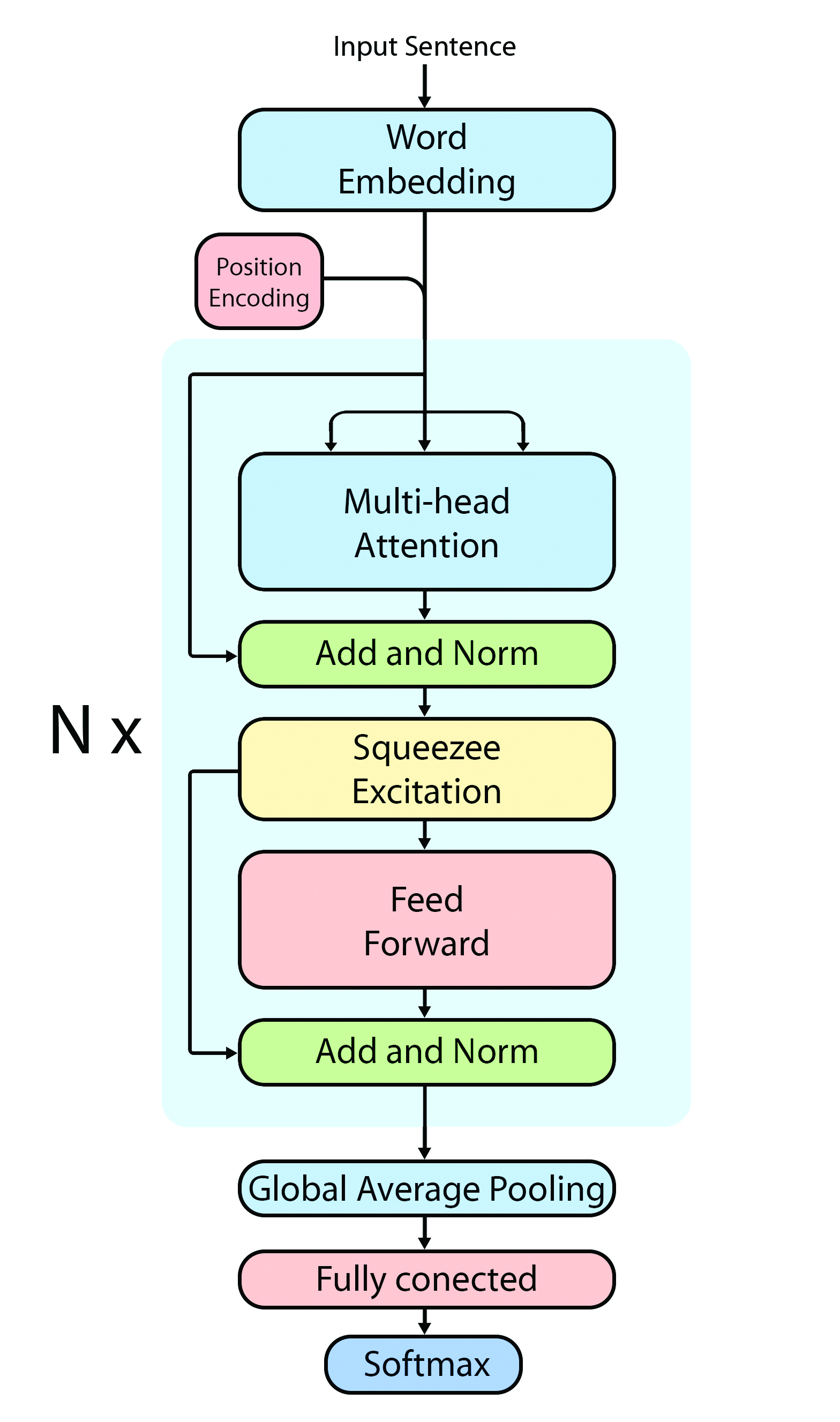}
\caption{Self-attention Neural Networks architecture for sentiments classification task.} \label{fig2}
\end{minipage}
\end{figure}

We proposed a simple model using a single modified $12$ heads Self-attention block (See Fig~\ref{fig2}), described below.

Original Sinusoidal Position Encoding \cite{vaswani2017} used ``adding" operation to incorporate positional informations as a input. That means while performing Self-attention, representation informations(Word Embeddings) and positional informations(Positional Embeddings) have the same weights (these two information are equal).

\begin{equation}
z = Embedding + PE
\end{equation}

In Vietnamese, we assumed that the positional information has more contributions to create contextual semantics than representation informations. Therefore, we used ``concatenate" operation to incorporate positional informations. That made representation informations may have a different weights with positional informations during the transformation process.

\begin{equation}
z = Concat(Embedding, PE)
\end{equation}

We added a block inspired by paper \emph{``Squeeze-and-Excitation Networks"}, Hu et al. \cite{hu2017} for the average attention mechanism and the gating mechanism by stacking a GobalAveragePooling1D layer then forming a bottleneck with two fully-connected layers (see Fig.~\ref{fig1}). The first layer is dimensionality-reduction layer with reduction ratio $r$ (in our experiment default is $4$) with a non-linear activation and then the second layer is dimensionality-increasing layer to return the result to $d_{model}$ dimension also with a sigmoid activation function, which scale the feature value into range $[0,1]$. It means this layer computes how much a feature incorporates information to contextual semantics. We call this technique Embedding Feature Attention. 

\begin{equation}
y = \sigma(W_{fc_2}\delta(W_{fc_1}x))
\end{equation}
Where $x$ is input of block. $y$ is output of block. $\sigma$ is a non-linear activation function. $\delta$ is a non-linear activation function. $W_{fc_1}$,$W_{fc_2}$ are trainable matrices.

\section{Experiments} \label{S:Expriment}
We implemented from scratch some layers that are needed for this work, such as: Scaled-dot product Attention, Multihead Attention, Feed-Forward Network and re-trained word embeddings for Vietnamese spoken language.

All experiments were deployed on 26GB RAM, CPU Intel Xeon Processor E31220L v2, GPU Tesla K80 for 20 epochs, 64 of batch size for comparison and all neural network models used focal-crossentropy as the training loss.
\subsection{Datasets} \label{datasets}

There is no public dataset for electronics product reviews in Vietnamese. We had to crawl user reviews from several e-commerce websites, such as Tiki, Lazada, shopee, Sendo, Adayroi, Dienmayxanh, Thegioididong, fptshop, vatgia. Based on our purposes, we chose some data fields to collect and store. Some data samples are presented in Tab.~\ref{tab1} below.

\begin{table}
\begin{center}
\caption{Examples for crawled data from e-commerce websites.}\label{tab1}
\begin{tabular}{|p{1.82cm}|p{3cm}|p{1.3cm}|p{5cm}|p{0.8cm}|}
\hline
username & product name & category & review & rating\\
\hline
user 1& 
\begin{otherlanguage*}{vietnamese}Samsung Galaxy A8+\end{otherlanguage*}& 
\begin{otherlanguage*}{vietnamese}điện thoại\end{otherlanguage*}&
\begin{otherlanguage*}{vietnamese}Ytt5ya 5t55\end{otherlanguage*}&
1/5\\
\hline
user 2& 
\begin{otherlanguage*}{vietnamese}Philips E181\end{otherlanguage*}& 
\begin{otherlanguage*}{vietnamese}điện thoại\end{otherlanguage*}&
\begin{otherlanguage*}{vietnamese}đang chơi liên quân tự nhiên bị đơ đơ. rồi tự nhảy lung tung. Bị như vậy là do game hay do máy v mọi người.\end{otherlanguage*}&
1/5\\
\hline
user 3&
\begin{otherlanguage*}{vietnamese}Philips E181\end{otherlanguage*}& 
\begin{otherlanguage*}{vietnamese}điện thoại\end{otherlanguage*}&
\begin{otherlanguage*}{vietnamese}Đặt màu vàng đồng mà giao màu bạc\end{otherlanguage*}&
2/5\\
\hline
user 4& 
\begin{otherlanguage*}{vietnamese}Oppo f7\end{otherlanguage*}&
\begin{otherlanguage*}{vietnamese}điện thoại\end{otherlanguage*}&
\begin{otherlanguage*}{vietnamese}Oppo f7 đang có chương trình trả trước 0\% và trả góp 0\% đúng không ạ?\end{otherlanguage*}&
2/5\\
\hline
user 5&
\begin{otherlanguage*}{vietnamese}Philips E181\end{otherlanguage*}& 
\begin{otherlanguage*}{vietnamese}điện thoại\end{otherlanguage*}&
\begin{otherlanguage*}{vietnamese}Giá đó mà không có camera kép, Vivo V9 đẹp hơn.\end{otherlanguage*}& 
2/5\\
\hline
user 6& 
\begin{otherlanguage*}{vietnamese}Samsung Note 7\end{otherlanguage*}& 
\begin{otherlanguage*}{vietnamese}điện thoại\end{otherlanguage*}&
\begin{otherlanguage*}{vietnamese}Cho em hỏi máy m5c của em hay bị tắt nguồn là do sao ạ?\end{otherlanguage*}&
4/5\\
\hline
user 7&
\begin{otherlanguage*}{vietnamese}Nokia 230 Dual SIM \end{otherlanguage*}& 
\begin{otherlanguage*}{vietnamese}điện thoại\end{otherlanguage*}&
\begin{otherlanguage*}{vietnamese}điện thoại vs Máy dùng tốt\end{otherlanguage*}&
4/5\\
\hline
user 8& 
\begin{otherlanguage*}{vietnamese}Oppo f7\end{otherlanguage*}& 
\begin{otherlanguage*}{vietnamese}điện thoại\end{otherlanguage*}&
\begin{otherlanguage*}{vietnamese}cho em hỏi giá oppo F7 hiện tại bên mình là bao nhiêu ạ?\end{otherlanguage*}& 
5/5\\
\hline
user 9& 
\begin{otherlanguage*}{vietnamese}Samsung Note 7\end{otherlanguage*}& 
\begin{otherlanguage*}{vietnamese}điện thoại\end{otherlanguage*}&
\begin{otherlanguage*}{vietnamese}Có màu đen ko vậy?\end{otherlanguage*}&
5/5\\
\hline
\end{tabular}
\end{center}
\end{table}

\begin{figure}[!htbp]
\begin{center}
\includegraphics[width=\textwidth]{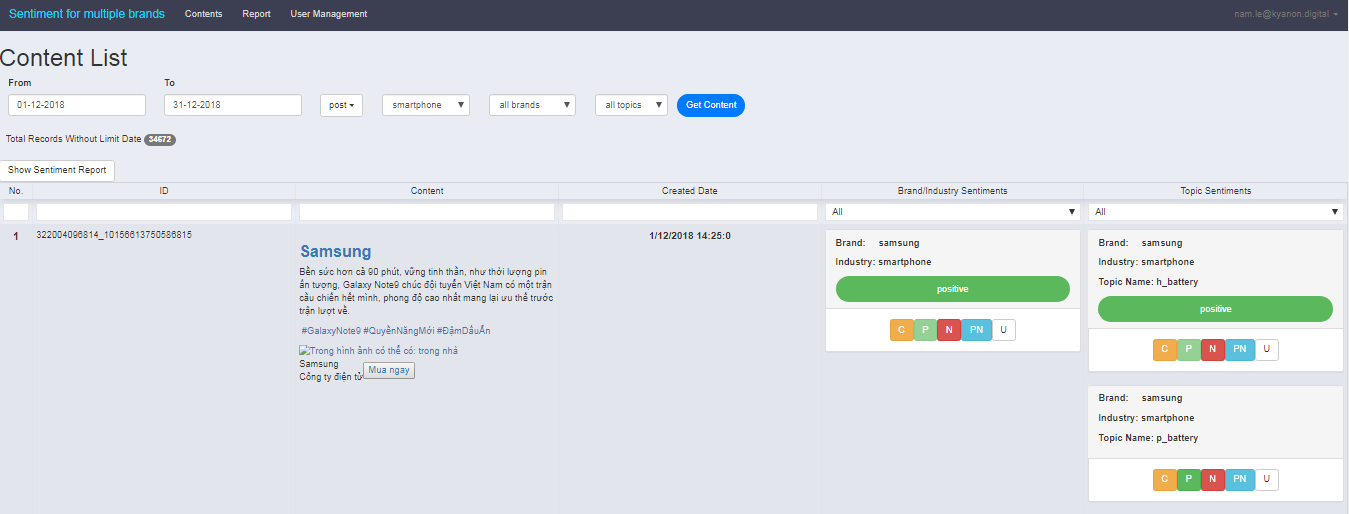}
\caption{Sentiment checking tool interface.}\label{fig3}
\end{center}
\end{figure}

After analyzing and visualizing, we found that the dataset was very imbalanced (see the description below) and noisy. There were some meaningless reviews ({\em{user1}} in Tab.~\ref{tab1}). Some of them did not have sentiments ({\em{user4}}, {\em{user8}} and {\em{user9}} in Tab.~\ref{tab1}). Sometimes, the ratings do not reflect the sentiment of reviews, (see {\em{user6}} in Tab.~\ref{tab1}). Therefore, a manual inspection step was applied to clean and label the data. We also built a tool for labeling process to made it smoothly and faster (see Fig.~\ref{fig3}).

\begin{description}
\item - Corpus have only 2 labels (positive and negative).
\item - Total 32,953 documents in labeled corpus:
\subitem Positives: 22,335 documents.
\subitem Negatives: 10,618 documents.
\end{description}

Next, to make the dataset balanced, we duplicated some short negative documents and segmented the longer ones. In final result we have over $43,500$ documents in corpus with $22,335$ positives and $21,236$ negatives.

Using for training models, we splitted corpus into 3 sets as following: training set: $27,489$, validation set: $6,873$, test set: $8,591$.
\subsection{Preprocessing}
For automatic preprocessing, we mainly used available researches. We applied a sentence tokenizer\cite{nltk} for each documents. All links, phone numbers and email addresses were replaced by {\em{``urlObj"}}, {\em{``phonenumObj"}} and {\em{``mailObj"}}, respectively. Words tokenizer from Underthesea\cite{underthesea} for Vietnamese was also applied.
\subsection{Embeddings}
We used fastText\cite{bojanowski2016} model for word embeddings. In many cases, users may type a wrong word accidentally or intentionally. fastText deals with this problem very well by encoding at the characters level. When users type wrong or very rare words or out-of-vocabulary words, fastText still can represent those words with an embedding vector that most similar to word met in trained sentences. This has made fastText become the best candidate to represent user inputs.

There had been no fastText pre-trained model for Vietnamese spoken language. Therefore, we trained fastText model for Vietnamese vocabulary as embedding pre-trained weights from a corpus over $70,000$ documents of multi-products reviews crawled from ecomerce sites mentioned in subsection~\ref{datasets} with no label. Rare words that occur less than $5$ times in the vocabulary were removed. Embedding size was $384$. After training, we had $5,534$ vocabularies in total.

\subsection{Evaluation results}
We used the same word embeddings as mentioned above for all models and evaluated all models on test set which has $8591$ documents. To demonstrate the significance of our model, we compare our model with $6$ base line RNNs models such as Long-Short Term Memory (LSTM), Gated Recurrent Units (GRU), bidirectional LSTM, bidirectional GRU, stacked bidirectional LSTM and stacked bidirectional GRU with the following configurations.

\begin{description}
\item - Vanilla LSTM and GRU: 1 layer with 1,024 units.
\item - Bidirectional model of LSTM and GRU: 1 layer with 1,024 units in forward and 1,024 units in backward.
\item - Stacked bidirectional model of LSTM and GRU: 2 stacked layers with 1,024 units in forward and 1,024 units in backward for each layer.
\end{description}

\begin{figure}[!htbp]
\begin{center}
\includegraphics[width=\textwidth]{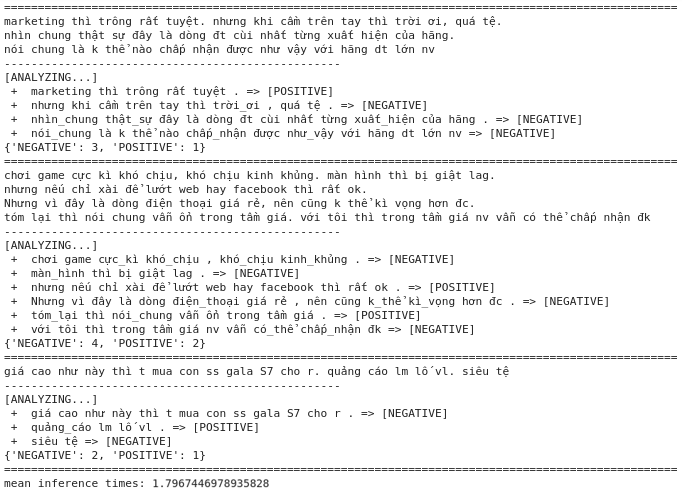}
\caption{Stacked bidirectional Long-Short term memory for Sentiments Analysis in Vietnamese examples} \label{fig4}
\includegraphics[width=\textwidth]{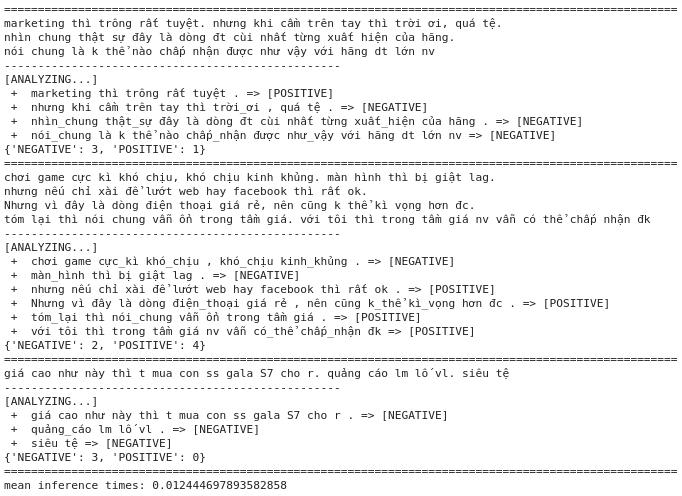}
\caption{Self-attention Neural Network for Sentiments Analysis in Vietnamese examples} \label{fig5}
\end{center}
\end{figure}

Table~\ref{tab2} shows that our model gave the best inference time with top accuracy in test set. Also, in fact, this model ran in prodution have shown good prediction than the top of baseline models, stacked Bidirectional Long-short Term Memory, especially with complex sentences such as ``\begin{otherlanguage*}{vietnamese}giá cao như này thì t mua con ss gala S7 cho r\end{otherlanguage*}", ``\begin{otherlanguage*}{vietnamese}quảng cáo lm lố vl\end{otherlanguage*}", ``\begin{otherlanguage*}{vietnamese}với tôi thì trong tầm giá nv vẫn có thể chấp nhận đk\end{otherlanguage*}" or ``\begin{otherlanguage*}{vietnamese}Nhưng vì đây là dòng điện thoại giá rẻ, nên cũng k thể kì vọng hơn đc.\end{otherlanguage*}" (See Fig.~\ref{fig4}, Fig.~\ref{fig5})

\begin{table}
\caption{Inference times and macro-f1 scores}\label{tab2}
\begin{center}
\begin{tabular}{||l|c|c||}
\hline
{\bfseries Methods}&{\bfseries Avg.inference time (s)}&{\bfseries Macro-f1 (\%)}\\
\hline
LSTM&0.4748&48.9(23)\\
bi-LSTM&0.9373&90.0(05)\\
stacked bi-LSTM&1.7967&90.1(32)\\
\hline
GRU&0.3738&48.9(23)\\
bi-GRU&0.5863&88.9(25)\\
stacked bi-GRU&1.4830&89.9(72)\\
\hline
{\bfseries Self-attention}&{\bfseries 0.0124}&{\bfseries 90.1(64)}\\
\hline
\end{tabular}
\end{center}
\end{table}

\section{Conclusion} \label{S:conclusion}
In this paper we demonstrated that using Self-attention Neural Network is faster than previous state of the art techniques with the best result in test set and achieved exceptionally good results when ran in prodution (Predictions make sense to human in unlabeled data with very fast inference time).

For future work, we plan to extend stacked multi-head self-attention architectures. We are also interested
in seeing the behaviour of the models explored in this work on much larger datasets (beyond the electronics product reviews) and more classes.

\section*{Acknowledgment} We thank our teammates, Tran A. Sang, Cao T. Thanh, and Ha H. Huy for helpful discussions and supports. 

%
%
%

\begin{thebibliography}{1}


\bibitem{hochreiter1997}
Sepp Hochreiter and Jurgen Schmidhuber. 
\newblock Long short-term memory.
\newblock In {\em Neural Computation, 9(8):1735–1780}, 1997.

\bibitem{schuster1997}
Mike Schuster and Kuldip K Paliwal.
\newblock{Bidirectional recurrent neural networks.}
\newblock{In {\em IEEE Transactions on Signal Processing, 45(11):2673–2681}, 1997}.

\bibitem{schneider1998}
Werner X. Schneider.
\newblock An Introduction to “Mechanisms of Visual Attention:A Cognitive Neuroscience Perspective.
\newblock {\em{URL: https://pdfs.semanticscholar.org/b719/918bdf2e71571a3cbb2a6aaaec3f1b6af9e6.pdf.}}, 1998.

\bibitem{esuli2006}
Andrea Esuli and Fabrizio Sebastiani. 
\newblock{Senti-wordnet: A publicly available lexical resource for opinion mining.}
\newblock{In {\em Proceedings of LREC, volume 6, pages 417–422}, 2006.}

\bibitem{baccianella2010}
Stefano Baccianella, Andrea Esuli, and Fabrizio Sebastiani. 
\newblock{Sentiwordnet 3.0: An enhanced lexical resource for sentiment analysis and opinion mining.}
\newblock{In {\em Proceedings of LREC, volume 10, pages 2200–2204}, 2010.}


\bibitem{mohammad2013}
Saif M. Mohammad, Svetlana Kiritchenko, and Xiaodan Zhu. 
\newblock{Nrc-canada: Building the state-of-the-art in sentiment analysis of tweets.}
\newblock{In {\em Proceedings of SemEval-2013.}, 2013.}

\bibitem{mnih2014}
Volodymyr Mnih et al.
\newblock Recurrent Models of Visual Attention.
\newblock In {\em Neural Information Processing Systems Conference (NIPS), 2014}.
\newblock {\em arXiv preprint arXiv:1406.6247}, 2014.

\bibitem{bahdanau2014}
Dzmitry Bahdanau, Kyunghyun Cho, Yoshua Bengio.
\newblock{Neural Machine Translation by Jointly Learning to Align and Translate.}
\newblock{accepted in {\em International Conference on Learning Representations (ICLR), 2015}.}
\newblock{{\em arXiv preprint arXiv:1409.0473 }, 2014.}


\bibitem{chung2014}
Junyoung Chung, Caglar Gulcehre, KyungHyun Cho, Yoshua Bengio
\newblock{Empirical Evaluation of Gated Recurrent Neural Networks on Sequence Modeling.}
\newblock{{\em arXiv preprint arXiv:1412.3555 }, 2014.}

\bibitem{cheng2016}
Jianpeng Cheng, Li Dong, and Mirella Lapata.
\newblock{Long short-term memory-networks for machine reading.}
\newblock{\em Computing Research Repository (CoRR), 2016.}
\newblock{{\em arXiv preprint arXiv:1601.06733 }, 2016.}

\bibitem{lu2016}
Jiasen Lu, Jianwei Yang, Dhruv Batra, and Devi Parikh
\newblock{Hierarchical question-image co-attention for visual question answering.}
\newblock{{\em Advances in Neural Information Processing Systems 29, pages 289–297, Curran Associates, Inc., 2016}}.

\bibitem{vo2016}
Duy Tin Vo and Yue Zhang.
\newblock Don’t Count, Predict! An Automatic Approach to Learning Sentiment Lexicons for Short Text.
\newblock {\em URL:https://www.aclweb.org/anthology/P16-2036}, 2016.

\bibitem{bojanowski2016}
Piotr Bojanowski, Edouard Grave, Armand Joulin, Tomas Mikolov
\newblock Enriching Word Vectors with Subword Information.
\newblock {\em arXiv preprint arXiv:1607.04606}, 2016.

\bibitem{kokkinos2017}
Filippos Kokkinos and Alexandros Potamianos.
\newblock{Structural attention neural networks for improved sentiment analysis.}
\newblock{{\em arXiv preprint arXiv:1701.01811}, 2017}.

\bibitem{daniluk2017}
Michal Daniluk, Tim Rocktaschel, Johannes Welbl and Sebastian Riedel.
\newblock{Frustratingly short attention spans in neural language modeling.}
\newblock{{\em arXiv preprint arXiv:1702.04521}, 2017.}

\bibitem{vaswani2017}
Ashish Vaswani, Noam Shazeer, Niki Parmar, Jakob Uszkoreit, Llion Jones, Aidan N Gomez, Lukasz Kaiser, and Illia Polosukhin. 
\newblock{Attention is all you need.}
\newblock{In {\em I. Guyon, U. V. Luxburg, S. Bengio, H. Wallach, R. Fergus, S. Vishwanathan, and R. Garnett, editors, Advances in Neural Information Processing Systems 30, pages 5998–6008, Curran Associates, Inc., 2017}.}
\newblock{{\em arXiv preprint arXiv:1706.03762}, 2017.}
\newblock{{\em URL:http://papers.nips.cc/paper/7181-attention-is-all-you-need.pdf}, 2017.}

\bibitem{gehring2017}
Jonas Gehring, Michael Auli, David Grangier, Denis Yarats, and Yann N Dauphin. 
\newblock{Convolutional sequence to sequence learning.}
\newblock{{\em arXiv preprint arXiv:1705.03122}, 2017.}

\bibitem{hu2017}
Jie Hu, Li Shen, Samuel Albanie, Gang Sun, Enhua Wu. 
\newblock{Squeeze-and-Excitation Networks.}
\newblock{{\em arXiv preprint arXiv:1709.01507}, 2017.}

\bibitem{zhou2018}
Yi Zhou, Junying Zhou, Lu Liu, Jiangtao Feng, Haoyuan Peng, and Xiaoqing Zheng.
\newblock{RNN-based sequence-preserved attention for dependency parsing.}
\newblock{{\em URL:https://www.aaai.org/ocs/index.php/AAAI/AAAI18/paper/view/17176 }, 2018.}

\bibitem{underthesea}
Vu Anh et al.
\newblock{Underthesea.}
\newblock{ULR: {\em{https://github.com/undertheseanlp/underthesea}}}.

\bibitem{nltk}
Natural Language Toolkit.
\newblock{URL: {\em{https://www.nltk.org/}}}.

\end{thebibliography}
%

\end{document}